\documentclass[11pt]{article}

\usepackage[english]{babel}

\usepackage[letterpaper,top=2cm,bottom=2cm,left=3cm,right=3cm,marginparwidth=1.75cm]{geometry}

\usepackage{amsmath}
\usepackage{xcolor}

\usepackage[authordate,
url=false,
doi=true]{biblatex-chicago}
\usepackage{csquotes}
\usepackage[english]{babel}
\usepackage{amssymb}
\usepackage{graphicx}
\usepackage{bm}
\usepackage{tabularx}
\usepackage{setspace}
\usepackage{array}
\usepackage[colorlinks=true, allcolors=blue]{hyperref}
\newcolumntype{C}{>{\centering\arraybackslash}X}
\title{Monitoring the calibration of probability forecasts with an application to concept drift detection involving image classification}
\author{Christopher T. Franck, Anne R. Driscoll,  Zoe Szajnfarber, William H. Woodall}

\begin{document}
\maketitle

\begin{abstract}
Machine learning approaches for image classification have led to impressive advances in that field. For example, convolutional neural networks are able to achieve remarkable image classification accuracy across a wide range of applications in industry, defense, and other areas. While these machine learning models boast impressive accuracy, a related concern is how to assess and maintain calibration in the predictions these models make. A classification model is said to be well calibrated if its predicted probabilities correspond with the rates events actually occur. While there are many available methods to assess machine learning calibration and recalibrate faulty predictions, less effort has been spent on developing approaches that continually monitor predictive models for potential loss of calibration as time passes. We propose a cumulative sum-based approach with dynamic limits that enable detection of miscalibration in both traditional process monitoring and concept drift applications. This enables early detection of operational context changes that impact image classification performance in the field. The proposed chart can be used broadly in any situation where the user needs to monitor probability predictions over time for potential lapses in calibration. Importantly, our method operates on probability predictions and event outcomes and does not require under-the-hood access to the machine learning model.

\end{abstract}

Keywords: image classification, calibration, cumulative sum (CUSUM) chart, dynamic control limits, probability prediction, statistical process monitoring, concept drift.

\section{Introduction}

Binary classification models such as logistic regression, tree-based approaches, and neural networks can furnish predictions that an event will occur in the probability space between zero and one. Perhaps surprisingly, these predictions may not actually reflect the empirical rate at which events occur, particularly when the model makes predictions outside of the training data. When predictions do not correspond to the actual event rate, we say a model is uncalibrated. Calibrated models provide predictions that correspond to the observed event rate. Thus, achieving and maintaining calibration is extremely important when model predictions are used to make decisions, as calibrated predictions help users understand the actual risk of prospective events. While calibration of prediction models has been well studied, there is a lack of methods available to prospectively monitor predictions to determine if and when predictive models lose calibration. \textcolor{black}{There are a variety of potential industrial and quality applications where a monitoring procedure that assesses calibration could be useful. Perhaps a machine learning model is used in a manufacturing setting to predict the probability that a part fails a downstream inspection, but a change in a material supplier compromises the ability for the model to predict probabilities properly. In the medical monitoring context, neural network-based approaches can predict surgical mortality (for example, \cite{Borner2024deep}), but such a model may lose calibration over time as new treatments appear and patient profiles change.} The purpose of this paper is to propose a calibration monitoring cumulative sum (CUSUM) approach that indicates whether and at what time a model becomes uncalibrated.

\begin{figure}
\centering
\includegraphics[width=1\linewidth]{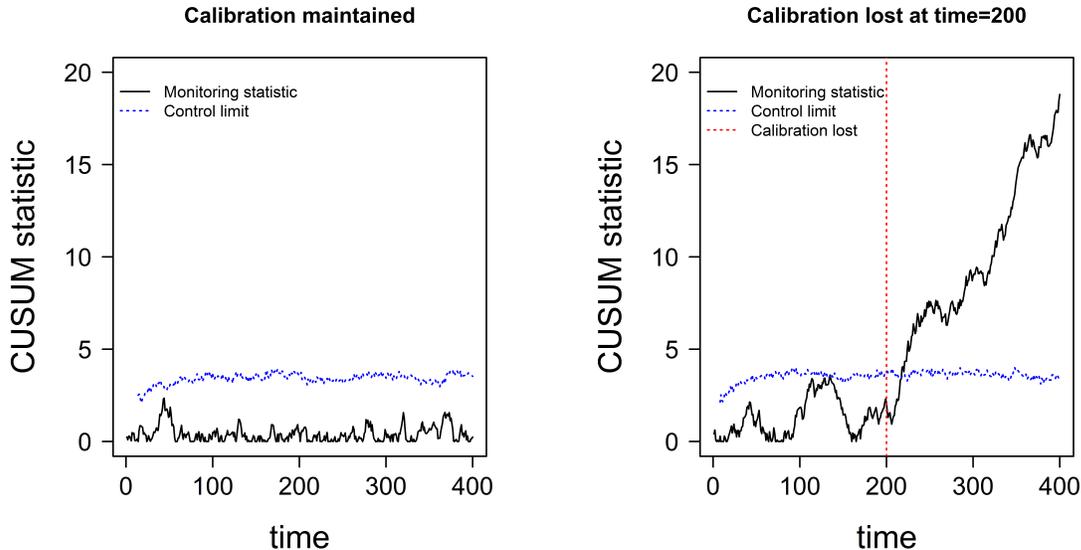}
\caption{\label{fig:fig1} Schematic of the proposed calibration CUSUM chart. The horizontal axis is time, and the vertical axis corresppnds to values of the CUSUM statistic. The solid black line corresponds to the observed value of the CUSUM statistic, and the dotted blue line provides the control limits. For the purpose of monitoring calibration of probability predictions, low values of the statistic correspond to the hypothesis that the model predictions are well calibrated while high values of the statistic correspond to a specific hypothesis that the predictions are not well calibrated. The left panel illustrates a scenario where calibration is maintained across the observed time course, as the CUSUM statistic does not exceed the control limits. The right panel includes a loss of calibration (vertical red line), and thus the CUSUM statistic (black) exceeds the control limits (blue) shortly thereafter, indicating to the user that probability predictions from the model are no longer well calibrated.}
\end{figure}

The core idea behind the proposed calibration CUSUM chart is illustrated in Figure \ref{fig:fig1}. The model's probability predictions are assessed at each time point (horizontal axis), and the calibration status of those predictions is summarized with our proposed CUSUM statistic (vertical axis) described in Section \ref{sec:CUSUMmethod}. In the left panel, the model continues producing calibrated predictions, thus the CUSUM statistic (solid black line) does not exceed the control limits (dotted blue line). In the right panel, calibration is lost at time 200, and the chart signals the loss of calibration shortly thereafter. Thus, in statistical process monitoring parlance, we refer to the status where predictions are well calibrated as in-control. This figure was generated using simulated data so the calibration status of the predictions is known. In practice the chart uses only predictions and event data and no knowledge about the true state of model calibration.

The cumulative sum (CUSUM) chart, first introduced by \textcite{Page1954_CUSUM}, is one of the essential monitoring tools within the broader field of statistical process monitoring. The chart can be designed to optimally detect small and persistent changes in the process \parencite{Montgomery2020}. While many different CUSUM charts have been developed, one popular method was deployed by \textcite{Steiner2000monitoring}. Here the authors developed a risk-adjusted CUSUM chart to monitor surgical outcomes while accounting for the variability in patient health status. Our proposed calibration CUSUM chart is a generalization of the approach presented in \textcite{Steiner2000monitoring}. \textcolor{black}{Specific details of the connection between the Steiner CUSUM chart and the proposed calibration CUSUM chart can be found in the Appendix.} We provide additional statistical process monitoring and concept drift detection overview in Section \ref{sec:cncptdrft_bgrd}.

In Section \ref{sec:case_study} we consider a case study where a convolutional neural network (CNN) is used as an image classification model to predict the probabilities that images contains vehicles. The analyst is concerned that as time passes the model may begin producing badly calibrated predictions, perhaps because new subtypes of vehicles that were not part of the original training sample appear in images in the field.

The remainder of this manuscript is organized as follows. In Section \ref{sec:cncptdrft_bgrd} we review concept drift and statistical process monitoring. Section \ref{sec:calib_bgrd} provides background in the assessment of calibration and techniques useful to recalibrate predictions. In Section \ref{sec:CUSUMmethod} we combine the concept of calibration with statistical process monitoring to develop the proposed calibration CUSUM chart. In Section \ref{sec:simstudy} we demonstrate the performance of the proposed calibration CUSUM chart using Monte Carlo simulation. In Section \ref{sec:case_study} we demonstrate the proposed calibration CUSUM method using an example involving the monitoring of CNN predictions for an image classification problem. Section \ref{sec:discussion} contains our concluding remarks.

\section{Concept drift and process monitoring background}
\label{sec:cncptdrft_bgrd}

\textcolor{black}{This section provides background on the statistical process monitoring concepts that are necessary for the development of our calibration CUSUM chart.} CUSUM control chart methods were first proposed by \textcite{Page1954_CUSUM} and later studied by \textcite{Hinkley1971inference}. They are used in quality control applications and for detecting concept drift in machine learning applications. The paper by \textcite{Page1954_CUSUM} is often cited in the quality control literature while \textcite{Hinkley1971inference} is not. The term “Page-Hinkley test”, however, is typically used in the concept drift detection literature. The Page-Hinkley method was included by \textcite{Klaiber2023most} as one of the ten most popular approaches for concept drift detection.

Consider the successive observations, $X_1, X_2, \ldots$ . We define $S_i = X_i - \mu_0 - K$, $i = 1, 2, \ldots$, where $K > 0$ and let $SS_t =\sum_{i=1}^t S_i$, $t = 1, 2, \ldots$. To detect an increase in the mean, a signal is given at time $T$ provided $U_T = SS_T - \text{min} (SS_i, i = 1, \ldots, T) > H$, where $H > 0$. This expression for the control chart statistic is used in the concept drift detection literature while in the quality control literature, we invariably find the mathematically equivalent expression $U_i = \text{max} (0, U_{i-1} + X_i - \mu_0 - K)$, $i = 1, 2, \ldots$, where typically $U_0 = 0$. A similar approach is used to detect a decrease in the mean. 

In quality control applications, the value of $\mu_0$ is typically either a target value or an estimate of the in-control mean based on a preliminary (Phase I) set of data. The value of $K$ is $(\mu_1 - \mu_0) / 2$, where $(\mu_1 - \mu_0)$ corresponds to the smallest size of the shift in the mean to be detected quickly. The value of $H$ is determined in order to control the in-control average run length (ARL), i.e., the average number of observations until a signal is given when in fact the mean of the process remains at $\mu_0$.

The CUSUM chart is based on a likelihood ratio approach and possesses an optimality property as discussed by \textcite[][pp. 138-139]{Hawkins2012cumulative}. Note that our proposed CUSUM chart is based on the likelihood ratio approach.

The approach taken with respect to the determination of the value $\mu_0$ is different in the concept drift detection literature. Most often the average of the current and all past observations is used, e.g. see \textcite{Gama2013evaluating} and \textcite{Goncalves2014comparative}. \textcite{Mouss2004test} wrote that $\mu_0$ is either “known before the jump or is estimated in a recursive way from the first available values”. In a similar way one incorporates prior data into parameter estimates with self-starting charts, but typically with a lag of one time unit. Using a lag seems more reasonable. \textcite{Wendler2021self} reviewed the literature on self-starting charts. 

If normality is assumed in quality control applications, then it is often written that $K = k\sigma$ and $H = h\sigma$, where $\sigma$ is the in-control standard deviation of the process. The statistical properties of the CUSUM would depend on the values of $k$ and $h$. It is written in the concept drift detection literature that normality of the input is required for the Page-Hinkley method to be used. \textcite{Mouss2004test} and \textcite{Agrahari2022concept}, for example, wrote that the method is used for the detection of an abrupt change of the average of a Gaussian signal. There is little to no discussion of the process variation in the concept drift detection literature. Although normality is often assumed to design CUSUM methods, CUSUM procedures have been developed for a great many continuous and discrete distributions.

Some concept drift detection methods are based on a window-limited approach where only the past $w (w>1)$ observations are used in the calculations. \textcite{Francis2025identifying} mentioned in passing the use of a moving window with the CUSUM method, but \textcite{Kronberg2024concept} reported that
\begin{quote}``Page-Hinkley does not employ sliding windows. Instead, each new seen instance is incorporated into (the overall average). To mitigate the influence of old observations and enhance responsiveness, a forgetting factor $\alpha$ is used to weight the observed values and the mean, favoring new observations.'' \end{quote} It is unclear how any forgetting factor would be used, although there is an implementation in the following software with a default value of $\alpha = 0.9999$: \href{https://scikit-multiflow.readthedocs.io/en/stable/api/api.html}{API Reference — scikit-multiflow 0.5.3 documentation}.

\textcite{Zhang2023concept} described the interface between concept drift and process monitoring. They proposed a general concept drift detection framework that is accomplished by temporally monitoring the score vector of the fitted model. While this approach is fruitful for parametric statistical models with a reasonable number of parameters, it is less practical for black box machine learning algorithms that may have hundreds, thousands, or more parameters which are not necessarily interpretable in the context of the problem, such as the case with the convolutional neural network model we use for image classification in our study. \textcite{Davis2020detection} used sliding windows and online stochastic gradient descent to monitor and update changes in logistic regression coefficients over time. This method currently appears to work in the logistic regression framework, and thus is also not applicable to the image classification problem we consider. To the best of our knowledge, there is no competing statistical process monitoring or concept drift detection approach that is (i) focused on monitoring the calibration of probability predictions and (ii) does not require under the hood access to the fitted model. \textcite{Zhang2023concept} requires score vectors to monitor highly parameterized models and \textcite{Davis2020detection} is implemented only within the class of logistic regression. Our method does not require access to the inner workings of the fitted model and is thus suitable for use on ML algorithms such as the neural net image classification example we present. This CUSUM method may be interpreted either in the context of process monitoring or, in contemporary machine learning terminology, as a tool for concept drift detection.

\section{Calibration background}
\label{sec:calib_bgrd}

\textcolor{black}{The calibration CUSUM chart we propose in Section \ref{sec:CUSUMmethod}} relies on (i) the ability to assess the calibration of a set of predictions relative to observed event data, (ii) the ability to restore calibration, i.e., ``recalibrate'' predictions through the use of an appropriate data-driven formula, and (iii) a sequential CUSUM approach that determines if and when the (potentially recalibrated) model predictions lose calibration at a future time point. We elaborate on (i) and (ii) here and (iii) in Section \ref{sec:CUSUMmethod}.

Suppose a model of interest $M$ has been trained on a set of relevant input data $Q_{\text{train}}$ and binary outcome data $y_{\text{train}}$. The model can now be used to output predictions. Let $M(Q) = x$ represent the model predictions for generic $Q$ (training or otherwise) where $0 \leq x \leq 1$ is the model's outputted probability-scale prediction. In our forthcoming example, the CNN is $M$, the input data $Q$ is an image that is being classified as an animal or a vehicle, $y=1$ if the image contains a vehicle and $y=0$ if the image contains an animal, and $x$ is the confidence score the CNN outputs in favor of the image containing a vehicle rather than an animal. The concern is that these $x$ values may be highly accurate (e.g., low misclassification rate, high area under a receiving operator characteristic curve), but the values $x$ may be uncalibrated and not correspond to the actual rates that vehicles appear in images.

We address (i)-(iii) with the  linear log odds (LLO) model used in \textcite{Guthrie2024boldness} and supporting software \parencite{Guthrie2024brcal}, because the LLO likelihood function is extremely flexible, effective, and convenient to implement in our proposed calibration CUSUM chart. \textcolor{black}{We have found that the LLO function is extremely flexible and able to characterize a wide variety of forms of miscalibration. For example, the simulation study in \textcite{Guthrie2024boldness} shows the ability of the LLO calibration framework to detect miscalibration that follows the Prelec function \parencite{Prelec1998probability} in addition to the LLO function.} The LLO function is as follows:

\begin{center}
\begin{equation}
\label{eq:llo}
g(x;\delta,\gamma)=\frac{\delta x^\gamma}{\delta x^\gamma + (1-x)^\gamma},
\end{equation}
\end{center}

\noindent
where $\delta$ is a log-odds shift parameter and $\gamma$ is a log-odds scale parameter. Equation (\ref{eq:llo}) is the mathematical expression that results from transforming $x$ to the log-odds scale, shifting the log-odds by $\delta$ and scaling by $\gamma$, then transforming back to the probability scale to obtain the adjusted forecast $g(x;\delta,\gamma)$. Note that if $\delta=\gamma=1$ then no adjustment occurs, i.e. $g(x;\delta=1,\gamma=1)=x$. In practice the recalibration step is usually performed using data separate from the training data so as to ensure the model produces well calibrated predictions for data it was not trained on.

Equation (\ref{eq:llo}) shows how a single probability prediction $x$ is recalibrated on the basis of parameters $\delta$ and $\gamma$ to obtain the adjusted forecast $g(x;\delta,\gamma)$. Statistical approaches to the calibration problem arise when a collection of such data are used to form a likelihood function that can be used for (i) calibration assessment, which can be achieved by likelihood ratio testing, (ii) recalibration, which basically amounts to maximum likelihood, and (iii) calibration CUSUM chart development (next section). Consider predictions $\mathbf{x} = <x_1,\ldots,x_n>^T$ and binary outcomes $\mathbf{y} = <y_1,\ldots,y_n>^T$. For independent outcome data stored in random vector $\mathbf{y}$, this leads to a Bernoulli likelihood with event probability $g(x;\delta,\gamma)$:

\begin{equation}\label{eq:likelihood}
    \pi(\mathbf{y} | \delta, \gamma) = \prod_{i=1}^n g(x_i;\delta, \gamma)^{y_i} \left[1-g(x_i;\delta, \gamma)\right]^{1-y_i}.
\end{equation}

We next illustrate the calibration assessment (i.e., testing) and recalibration using an image classification problem. We consider the CIFAR-10 data \parencite{Krizhevsky2009cifar}. While we describe this data more thoroughly in Section \ref{sec:case_study}, here we examine the calibration status of the 10,000 test set predictions from a CNN that was trained to classify the CIFAR 10 images as vehicles (events) versus animals (non-events).  CNNs produce confidence scores $x$ that are between zero and one but are not expected to necessarily be well calibrated as predicted probabilities, so this analysis shows how to recalibrate these confidence scores.

We conduct this analysis using the \textbf{BRcal} \parencite{Guthrie2024brcal} software package from CRAN, due to its convenient implementation of calibration methods. (Note we use \textbf{BRcal} for MLE recalibration and not boldness recalibration that was the main contribution of \textcite{Guthrie2024boldness}.) Maximizing the log likelihood function based on Equation (\ref{eq:likelihood}) is about as complex as logistic regression and thus can be implemented with standard optimization routines.

\begin{figure}
\centering
\includegraphics[width=1\linewidth]{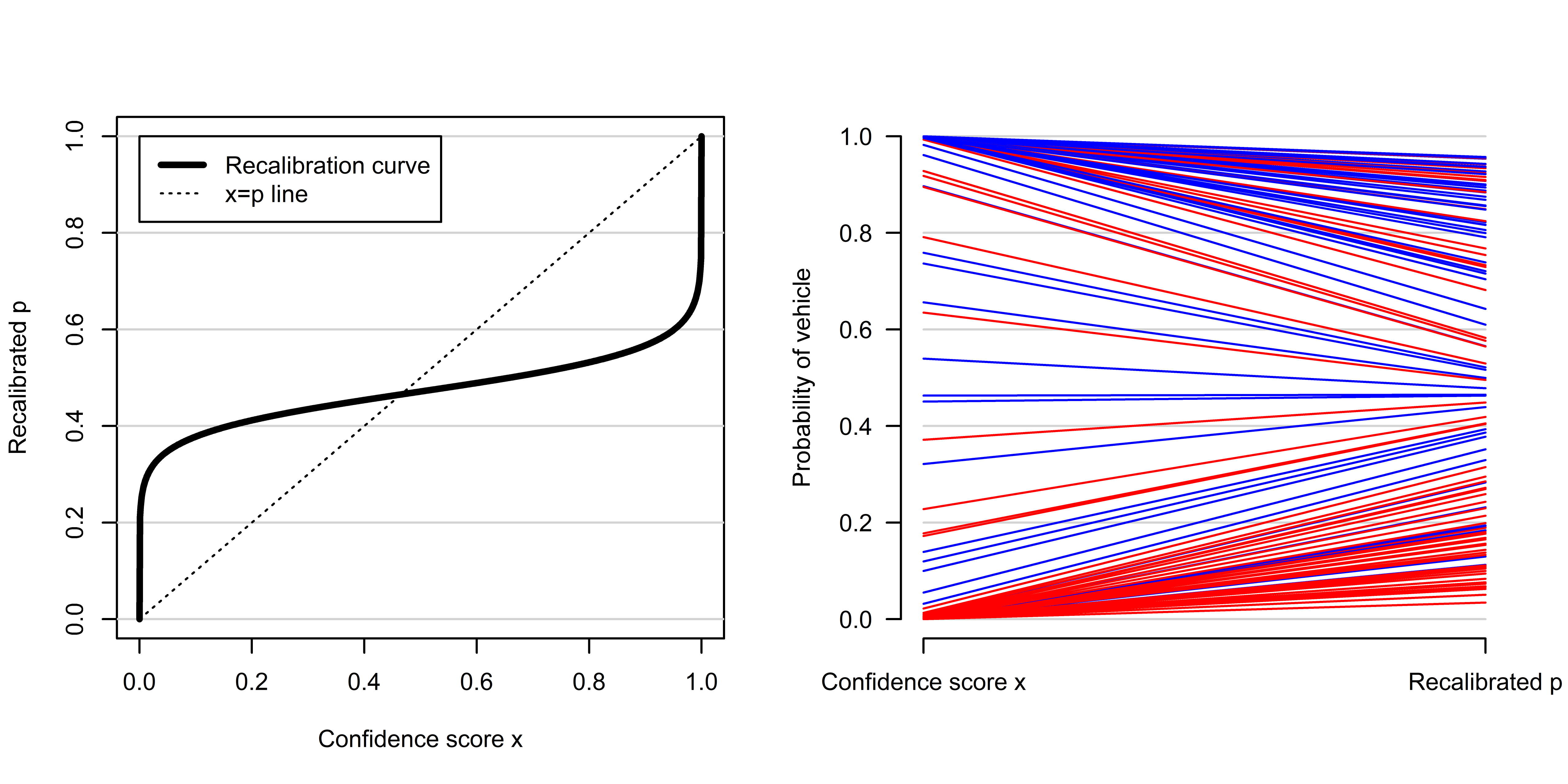}
\caption{\label{fig:fig3} Illustration of calibration analysis using CNN confidence scores $x$ and LLO recalibrated predictions $p$. The event being predicted is whether an image contains a vehicle. The left panel shows the recalibration curve. The solid black line shows the transformation needed for badly calibrated CNN confidence scores $x$ to be translated to well calibrated predictions $p$. The $x=y$ line would arise if the original $x$ predictions were well calibrated initially. The right panel is a line plot where each line corresponds to a prediction from a single image. Blue lines are predictions for images that actually contain a vehicle, and red lines are for images that contain animals. This panel shows that the initial $x$ predictions were overly-confident, and the recalibration function brings them further away from the extremes of zero and one. Maximum likelihood analysis was performed using the 10,000 predictions in the CIFAR-10 test set. MLEs for these data are $\hat{\delta}=0.89$ and $\hat{\gamma}=0.17$ The right panel line plot shows a smaller subset of the predictions for visual clarity.}
\end{figure}

Figure \ref{fig:fig3} contains two panels that summarize the calibration status and recalibration of CNN confidence scores $x$ and recalibrated predictions $p$ for the 10,000 images in the CIFAR-10 test set. \textcolor{black}{We note that a model is only calibrated if the data plausibly fall on the x=y line as this is the line that implies confidence scores correspond to the actual probability of events.} The CNN confidence scores $\mathbf{x}$ are badly calibrated. The left panel shows the recalibration curve, which is the LLO function in Equation (\ref{eq:llo}) with maximum likelihood estimates $\hat{\delta}=0.89$ and $\hat{\gamma}=0.17$ plugged in. When predictions are uncalibrated, maximizing the likelihood in (\ref{eq:likelihood}) provides the shift value $\hat{\delta}$ and scale value $\hat{\gamma}$ that bring the $\mathbf{x}$ predictions in line with the observed $\mathbf{y}$ outcomes. Then, the recalibrated prediction corresponding to the $i\text{th}$ observation $p_i$ is 

$$p_i = g(x_i;\hat{\delta},\hat{\gamma}).$$ 

The original predictions $x$ would be well calibrated if the recalibration curve followed the $x=p$ line, but obviously the observed function is quite different from the $p=x$ line. Unsurprisingly, a likelihood ratio test of the null hypothesis that these predictions are well calibrated (as implemented in \textcite{Guthrie2024boldness}) exhibits a $\text{p-value}<0.0001$.

The right panel of Figure \ref{fig:fig3} is a line plot that shows how the original CNN confidence scores $x$ are too close to the extremes of zero and one, and the LLO recalibration shifts them to less extreme values so as to achieve calibration. This analysis illustrates how contemporary machine learning approaches such as neural nets may require further attention in order to make their predictions well calibrated.

There are many reasons that model predictions may be uncalibrated or subsequently become uncalibrated. Certain types of models, such as CNNs, tend to not produce calibrated predictions outside of the data used for training without further recalibration, as shown above. This is because CNNs output confidence scores to express their predictions in classification problems. This is typically accomplished by allowing early layers of the CNN to operate on an unrestricted prediction space, then the last layer implements an activation function that constrains classification predictions for an object to (a) be between zero and one for all classes, and (b) make the class predictions sum to one. A commonly used example is the softmax function, which maps unconstrained predictions on the real line to the constraints (a) and (b) simply by exponentiating the unconstrained predictions then dividing each of these exponentiated predictions by the sum to obtain the what are referred to as confidence scores. These confidence scores satisfy the basic axioms of probability but are in no way guaranteed to yield predictions that are well calibrated.  More commonly, a model can initially be trained sufficiently well for accuracy and calibration, but the model may be deployed in situations where the conditions shift such that they no longer resemble the training conditions and calibration can be lost. Even when post training calibration is achieved, there are no broad guarantees that a model will continue furnishing well calibrated predictions, especially under evolving conditions that may differ from the training set.

\section{The calibration CUSUM chart}
\label{sec:CUSUMmethod}

In the previous section we demonstrated how to recalibrate the initially uncalibrated $x$ predictions to well calibrated $p$ predictions using available data analyzed at a single time point. The remainder of this paper focuses on determining whether and when these $p$ predictions may lose calibration over time. As the conditions in which new data are collected may change over time, there are unfortunately no guarantees that initially well calibrated $p$ predictions will continue to perform well indefinitely into the future. Even well calibrated predictions may eventually lose calibration if the model $M$ is used in conditions that differ from the training conditions.

\begin{figure}
\centering
\includegraphics[width=1\linewidth]{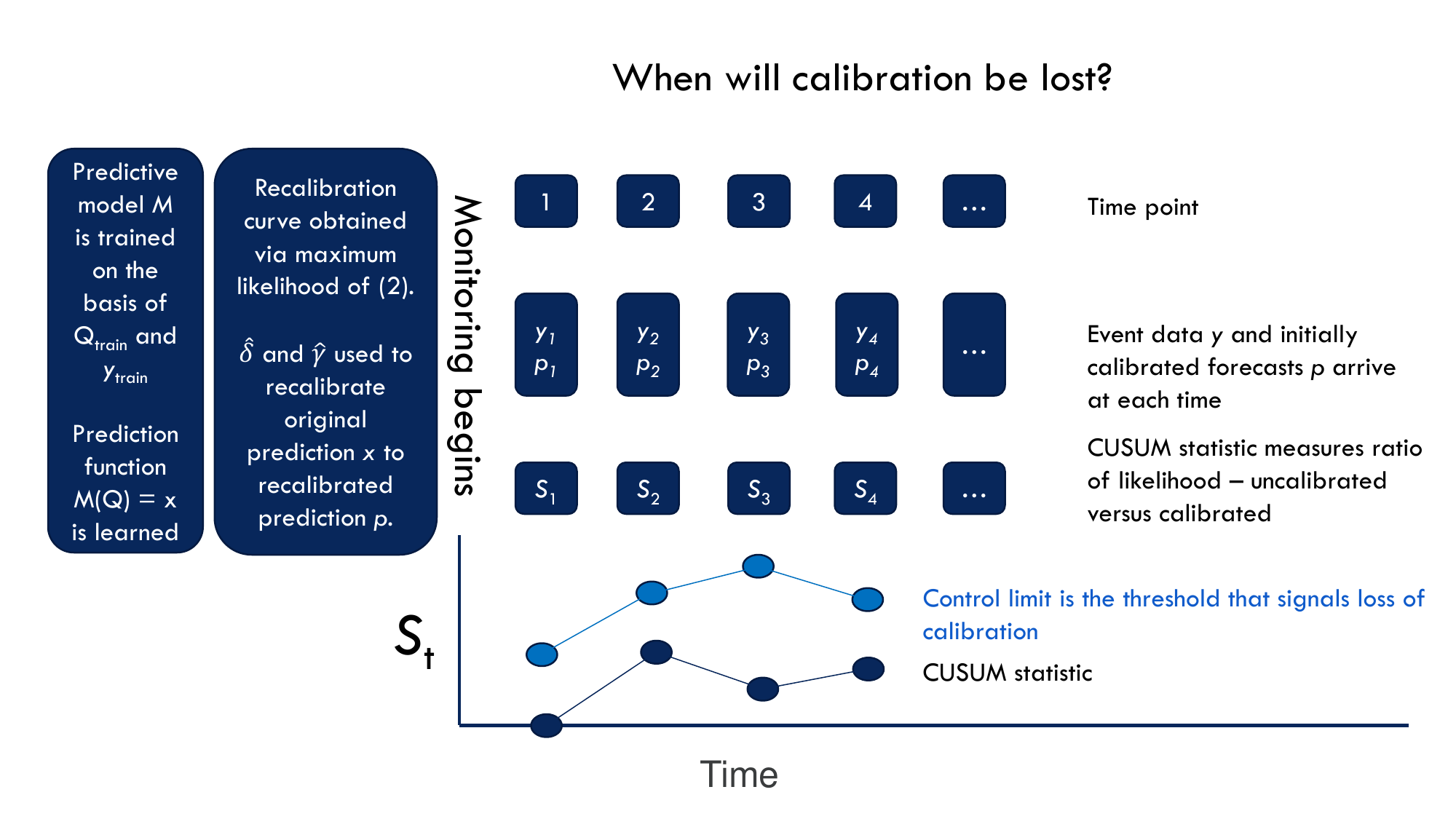}
\caption{\label{fig:fig2} Flowchart for how the calibration monitoring scheme works. Model is trained and initial calibration is verified, then monitoring begins. At each time point $t$, prediction $p_t$ and event data $y_t$ are collected, a CUSUM statistic $S_t$ is computed along with control limits. Monitoring continues until the chart signals. Full details of the calibration CUSUM monitoring approach are in Section \ref{sec:CUSUMmethod}.}
\end{figure}

Figure \ref{fig:fig2} provides a flowchart for how calibration monitoring is used in practice. The first two steps are to train model $M$ then assess whether $M$ is well calibrated, recalibrating if necessary. Calibration has been studied extensively, see \textcite{silvafilho2023classifier} for a broad review and \textcite{Guthrie2024boldness} for recent advances involving hypothesis testing, Bayesian model selection, and related emboldening strategies.

During the monitoring phase, event data $y_t$ and recalibrated predictions $p_t$ are collected, a CUSUM statistic is computed, adaptive control limits are computed, and these statistics are compared prospectively until such a time that the chart signals. Details follow.

We let $t=1,2,\ldots$. Let $n_t \geq 1$ be the number of independent trials at timepoint $t$. For example, a variety of sensors may generate an image at the same time point and all of these images would be simultaneously considered. \textcolor{black}{The value $i$ indexes the trials that happen within time $t$, where each trial is either an event $y_{it}=1$ or a non-event $y_{it}=0$.} Let $p_{it}$ be the probability prediction of trial $i$ at time $t$ for $i=1,\ldots,n_t$.  These $p_{it}$ values arise from the combined training and calibration assessment steps described in Section \ref{sec:calib_bgrd}.

First we consider the joint density of event data at time $t$ under the hypothesis that the model predictions are calibrated at time $t$. We let $\bm{y}_t = (y_{1t},\ldots,y_{n_t t})$ represent the set of trials at time $t$, and $\bm{p}_t= (p_{1t},\ldots,p_{n_t t)}$ the corresponding set of predictions. We define $f_c(y_{it})$ as the probability density function corresponding to $p_{it}$ being well calibrated for $y_{it}$. If $p_{it}$ is a well calibrated prediction for event $y_{it}$, then $y_{it}\sim \text{Bernoulli}(p_{it}$), and

$$
f_c(y_{it}) = p_{it}^{y_{it}} (1-p_{it})^{1-y_{it}}.
$$

The joint density assuming calibration at time $t$ is

$$
f_c(\bm{y}_t)= \prod_{i=1}^{n_t} f_c(y_{it}) = \prod_{i=1}^{n_t} p_{it}^{y_{it}} (1-p_{it})^{1-y_{it}}.
$$

Next we consider the joint density of event data $ \bm{y}_t$ at time $t$ under the hypothesis that the model predictions $\bm{p}_t$ are \textit{not} calibrated. Define $f_u(y_{it}; \delta, \gamma)$ as the joint probability density function corresponding to adjusted forecasts $g(p_{it};\delta,\gamma)$. Under this model $y_{it} \sim \text{Bernoulli}(g(p_{it};\delta,\gamma))$ and 

\begin{equation}
\label{eq:uncal_dens}
f_u(\bm{y}_t;\delta,\gamma) = \prod_{i=1}^{n_t} g(p_{it};\delta,\gamma)^{y_{it}}[1-g(p_{it};\delta,\gamma)]^{1-y_{it}}.
\end{equation}

Equation (\ref{eq:uncal_dens}) reflects the possibility that at time $t$ the originally well calibrated model forecasts $p_{it}$ are no longer well calibrated at time $t$, but rather depart from calibration according to the LLO function. Thus within Equation (\ref{eq:uncal_dens}) \textcolor{black}{we are considering values of $\delta$ and $\gamma$ are no longer equal to one. We use two one-sided charts to monitor each parameter}. 

Like other statistical process monitoring approaches, the calibration CUSUM chart requires the user to specify specific out-of-control values for the purpose of detecting a change in the process, The specific alternative values to be used in the CUSUM chart are $\delta_a$ and $\gamma_a$. The values $\delta_a$ and $\gamma_a$ describe a shift and scale on the log odds scale and should be chosen as the magnitude of departure that would be important to detect quickly. \textcolor{black}{Previous literature \parencite{Steiner2000monitoring} has used a doubling or halving of the odds ratio as a meaningful change. However, this choice is inherently subject-specific and may involve obtaining input from domain experts.}

The $\delta_a$ and $\gamma_a$ values are incorporated into the log likelihood expression used in the CUSUM statistics. Four specific charts that are fairly standard are a ``shift down'' chart where $\delta_a<1$ indicating that the observed event rate is lower in terms of log odds than the original predictions, a ``shift up'' chart where $\delta_a>1$ indicating an event rate with higher log odds than the original predictions, a ``scale down'' where $\gamma_a<1$ indicating events are less variable than the original predictions suggest, and a ``scale up'' chart where $\gamma_a>1$ indicating event rates are more variable than the original predictions suggest. \textcolor{black}{We have chosen to present univariate charts rather than bivariate charts in order to facilitate these convenient interpretations.}

The log-likelihood ratio $W_t$ between $f_c(\bm{y}_t)$ and $f_u(\bm{y}_t; \delta_a, \gamma_a)$ quantifies the competing hypotheses that the forecasts $\bm{p}_t$ are well calibrated versus not well calibrated:

\begin{equation}
\label{logratio}
W_t = \text{log}\Big[ \frac{f_u(\bm{y}_t;\delta_a,\gamma_a)}{f_c(\bm{y}_t)} \Big].
\end{equation}

The resulting CUSUM monitoring statistics are

\begin{equation}
S_t = \text{max}(0,S_{t-1} + W_t),\text{ } t=1,2,\ldots
\end{equation}

At time $t=1$ the statistic $S_{t-1}=0.$ Following Figure \ref{fig:fig1}, event data $\bm{y}_t$ and predictions $\bm{p}_t$ become available at time $t$, then $W_t$ and $S_t$ are computed, and the CUSUM statistic $S_t$ is monitored. This approach was devised on the basis of the literature in prediction calibration but is coincidentally also a generalization of the approach in  \textcite{Steiner2000monitoring}, which corresponds to setting $\gamma=1$. 

\subsection{Control limits}
\label{subsec:CL}

Control limits are needed for implementing the CUSUM procedure effectively. Dynamic control limits that maintain a constant conditional false alarm rate (CFAR) over time, typically determined using simulation, are recommended. These limits are commonly referred to in the literature as dynamic probability control limits (DPCLs). The CFAR-based approach has proven especially effective in control chart design when covariates or risk factors vary over time.
\textcite{ZhangWoodall2015_DPCL} proposed DPCLs for the risk-adjusted CUSUM chart introduced by \textcite{Steiner2000monitoring}, accommodating time-varying risk factors. Building on this, \textcite{ AytacogluWoodall2020_DPCL} applied the CFAR framework to design control limits for monitoring proportions using a CUSUM chart when sample sizes vary over time. Further advances were made by \textcite{Aytacoglu2022_MEWMA,Aytacoglu2023_AEWMA}  who extended the DPCL methodology to adaptive EWMA and multivariate exponentially weighted moving average (MEWMA) control charts, respectively. 

\textcite{Haq2024_DPCL_AMEWMA} further developed DCPLs for the adaptive MEWMA chart, while \textcite{HaqWoodall2023_EstErrorCFAR} examined how the estimation error impacts the CFAR when control limits are based on estimated DPCLs. Most recently, \textcite{Zhang2025_WeibullWACUSUM} expanded the application of DPCLs to weighted adaptive CUSUM charts for Weibull-distributed data. For a broader overview of CFAR applications in control chart design, see \textcite{Driscoll2021_CFAR}.

For the proposed calibrated CUSUM charts, we develop a DPCL-based approach as described in Section \ref{subsec:CL}. The algorithm to produce the DPCLs for the calibrated CUSUM chart (visualized as a blue line in Figure \ref{fig:fig1}) is as follows. We presume the user has selected relevant out-of-control values for $\delta_a$ and $\gamma_a$ such that they are not both equal to 1. We also presume the user has chosen $\alpha$ as the CFAR which will correspond to the upper quantile of the sampling distribution of the CUSUM statistics used to form the control limits. We proceed with the following steps: 

\begin{itemize}
\item Set number of Monte Carlo draws $M$ to a large value, such as 5,000.
\item At time $t$, draw $M$ sets of Bernoulli draws, i.e., $y_{it}^{(q)} \sim \text{Bernoulli}(p_{it})$ for $i=1,\ldots,n_t$ and $q=1,\ldots,M$. Note that the $y_{it}^{(q)}$ draws are calibrated with respect to probability forecasts $p_{it}$ by their construction.
\item Define $g_{it} = g(p_{it},\delta_a, \gamma_a)$. This $g_{it}$ is the adjusted forecast for this chart. We now have $n_t$ \textcolor{black}{$p_{it}$} values, $n_t$ $g_{it}$ values, and $M$ $\bm{y}_t^{(q)}$ sets of simulated trials. Denote the set of adjusted forecasts at time $t$ as $\bm{g}_t$.
\item Compute the log-likelihood ratios involving $\bm{x}_t$, $\bm{g}_t$, and $\bm{y}_t^{(q)}$ values:\\
$W_t^{(q)} = \text{log}\Big[ \frac{f_u(\bm{y}_t^{(q)})}{f_c(\bm{y}_t^{(q)};\delta_a,\gamma_a)} \Big]$ for $q=1,\ldots,M$.
\item Obtain $S_t^{(q)} = \text{max}(0,S_{t-1}^{(q)} +W_t^{(q)})$ for $q=1,\ldots,M$. Note that at time $t=1$ the value $S_{t-1}^{(q)}=0$. For $t>1$ the $S_{t-1}^{(q)}$ values are drawn with replacement from the collection of $S_{t-1}^{(q)}$ statistics below the $(1-\alpha)$ upper quantile of the previous step. 
\item The control limit at time $t$ is the $(1-\alpha)$ upper quantile of $S_t^{(q)}$. In other words, if the observed $S_t$ statistic is above the upper $(1-\alpha)$ quantile of the $S_t^{(q)}$ distribution, the chart signals.
\end{itemize}


There are a variety of remedies to consider if and when the calibration CUSUM chart signals a loss of calibration. One solution is to simply use recently gathered data to perform another recalibration, then continue with monitoring as before. A better solution is probably going to involve consultation with the domain experts in the subject area where monitoring is taking place in order to assess questions such as: Did the models being used in the field get updated or changed but not initially recalibrated? Have the conditions (e.g., weather, season, time of day) under which monitoring is occurring different from the training set? Have new subclasses of objects appeared in the field which were not part of the training sample? We consider this last possibility in the case study in Section \ref{sec:case_study}.
 
\section{Simulation study}
\label{sec:simstudy}

We conducted a Monte Carlo simulation study to evaluate the performance of the calibration CUSUM chart across a range of settings. Specifically, we examined the zero-state ARL, standard deviation of the run length (SDRL), and various quantiles (10th, 25th, 50th, 75th, 90th). For all simulations, the number of Monte Carlo draws $M=5,000$ was used to characterize the null distribution of the CUSUM statistic, and the upper tail threshold was set to $\alpha=0.005$. We chose a higher, less stringent $\alpha$ value for the simulation study compared with the case study in the next section to shorten run lengths and thus reduce the computational burden. Predictions $p$ were drawn from a uniform(0,1) distribution. 

Due to the high computational cost of obtaining the DPCLs, we generated a single vector $p$ per setting along with the corresponding DPCLs. On this basis of this $p$ vector, we sampled 10,000 Monte Carlo draws of $y$. Depending on the calibration scenario, $y$ was simulated either directly from $p$ (i.e., the well calibrated case), or $y$ was generated from LLO shifted and/or scaled $p$ (i.e., the uncalibrated case). More details on the specific generation of $y$ for both well calibrated and the uncalibrated case can be found below.

For each simulation scenario, we computed the ARL, SDRL, and quantiles of the run length based on the available $p$ predictions. To explore the extent to which the number of observations per time point might affect the performance of the chart, we considered four specific cases. The ``Fixed $n=1$'' and ``Fixed $n=3$'' settings had one and three observations per time point, respectively. For the ``lPoisson $\lambda=1$'' and ``lPoisson $\lambda=3$'' we drew the number of observations from a Poisson distribution with mean 1 and 3 respectively, then added one to this draw (``l'' stands for location shifted) so as to always have at least one observation per time point. We explored both well calibrated and uncalibrated setting using these methods.

Specific to the well calibrated settings, we generated the $y$ binary outcomes directly from the $p$ predictions, that is each $y \sim \text{Bernoulli}(p)$. Since the events follow the predictions exactly, these predictions are well calibrated by construction. The well calibrated setting is the ``in control'' setting.  We considered ``shift up'' and ``scale up'' alternatives as previously defined for parameters $\delta_a$ and $\gamma_a$. These indicate the snallest departure from calibration that the calibration CUSUM chart is meant to detect quickly. Finally, we report the relevant values arising from a geometric distribution with $\text{probability}=1/200$, as this explains the theoretical wait times for the first event to happen at $\alpha=0.005$ given that these charts are well calibrated and thus ``in control'' by their construction.

Specific to the uncalibrated setting, predictions $p$ were again generated from a uniform(0,1) distribution. However, the log odds of $p$ were shifted by $\delta$ and/or scaled by $\gamma$ prior to the generation of $y$. Thus, the $y$ outcomes do not arise at the original rate $p$ and the predictions are uncalibrated. We achieved miscalibration using $\delta$ and $\gamma$ values of 1/2 and 2 in this study. Aside from miscalibrating data to assess the calibration CUSUM chart's ability to detect lack of calibration, the rest of the simulation study is similar to what has been described.

Table \ref{tab:IC} shows the result for our well calibrated settings. Across all alternative specifications of $\delta_a$ and $\gamma_a$ and all distributional settings on the number of events per time point, the run length properties of the calibration CUSUM chart closely resemble the geometric distribution mean, standard deviation, and quantiles with false signal rate $\alpha$. We expect a geometrically distributed in-control run length distribution when we use DPCLs that control the CFAR. The distribution is not exactly geometric and this is attributed to the discreteness of the data. This analysis demonstrates the statistical properties of the calibration CUSUM chart hold when the predictions $p$ being monitored are truly well calibrated. 

Table \ref{tab:OOC} shows the run length properties of the calibration CUSUM chart when the predictions $p$ are not calibrated for event data $y$. In all cases, the chart signals much sooner than the well calibrated settings in Table \ref{tab:IC}. This observation holds for all alterative parameters and distributions of events per time point. These results demonstrate the ability of the calibration CUSUM chart to detect departures from calibration effectively. Taken together, the simulation results presented in Tables \ref{tab:IC} and \ref{tab:OOC} show the ability of the calibration CUSUM chart to signal only at the prescribed false alarm rate $\alpha$ while predictions are calibrated, but is able to signal rapidly when predictions are uncalibrated.

\begin{table}
  \centering
  \begin{tabular}{lcccccccc}
    \hline
    \textbf{Alternatives} & \bf{Distribution} & \bf{ARL} & \bf{SDRL} & 10th & 25th & 50th & 75th & 90th \\
    \hline
    $\delta_a=2,\gamma_a=1$ & Fixed $n=1$ & 233.58 & 209.55 &  41.00 &  84.00 & 169.00 & 319.00 & 507.00  \\
    $\delta_a=1,\gamma_a=2$ & Fixed $n=1$ & 225.27 & 212.32 &  40.00 &  73.00 & 163.00 & 302.00 & 497.00 \\ \hline
    $\delta_a=2,\gamma_a=1$ & Fixed $n=3$ & 212.12 & 203.03 &  29.00 &  67.00 & 152.00 & 295.00 & 469.00 \\
    $\delta_a=1,\gamma_a=2$ & Fixed $n=3$ & 207.49 & 199.28 &  26.00 &  64.00 & 149.00 & 286.00 & 470.20 \\ \hline
    $\delta_a=2,\gamma_a=1$ & lPoisson $\lambda=1$ & 208.97 & 202.02 &  29.00 &  64.00 & 149.00 & 286.00 & 475.00 \\
    $\delta_a=1,\gamma_a=2$ & lPoisson $\lambda=1$ & 208.35 & 196.38 &  29.00 &  69.00 & 145.00 & 288.00 & 464.00 \\ \hline
    $\delta_a=2,\gamma_a=1$ & lPoisson $\lambda=3$ & 209.77 & 199.99 &  30.00 &  65.00 & 150.00 & 296.00 & 471.00 \\
    $\delta_a=1,\gamma_a=2$ & lPoisson $\lambda=3$ & 205.38 & 195.21 &  29.00 &  68.00 & 146.00 & 280.00 & 459.00 \\ \hline
     & $\text{Geometric}(\alpha=1/200)$ & 200 & 199.5 & 21 & 57 & 138 & 276 & 459\\ \hline
 \end{tabular}
 \caption{Monte Carlo simulation study results for run length in the well calibrated settlings, i.e., $\delta=\gamma=1$. First column shows values the ``shift up'' and ``scale up'' alternatives.  The second column describes the distribution of possible events per time point. lPoisson refers to a Poisson random variable shifted up by 1 to guarantee at least one observation per time point. Remaining columns provide average, standard deviation, and various quantiles of run length.}
  \label{tab:IC}
\end{table}

\begin{table}
  \centering
  \begin{tabular}{lcccccccc}
    \hline
    \textbf{Parameters} & \bf{Distribution} & \bf{ARL} & \bf{SDRL} & 10th & 25th & 50th & 75th & 90th \\
    \hline
    $\delta=2,\gamma=1$ & Fixed $n=1$ & 36.77 & 29.21 &  8.00 & 15.00 & 29.00 & 50.00 & 76.10 \\
    $\delta=2,\gamma=1$ & Fixed $n=3$ & 19.51 & 12.66 &  8.00 & 10.00 & 15.00 & 25.00 & 36.00 \\
    $\delta=2,\gamma=1$ & lPoisson $\lambda=1$ & 30.59 & 14.69 & 21.00 & 21.00 & 24.00 & 35.00 & 49.00 \\
    $\delta=2,\gamma=1$ & lPoisson $\lambda=3$ & 23.23 & 7.88 & 19.00 & 19.00 & 19.00 & 24.00 & 33.00 \\ \hline
    $\delta=1,\gamma=2$ & Fixed $n=1$ & 43.11 & 21.74 & 26.00 & 26.00 & 35.00 & 51.00 & 72.00 \\
    $\delta=1,\gamma=2$ & Fixed $n=3$ & 18.01 & 9.18 & 10.00 & 11.00 & 15.00 & 22.00 & 31.00 \\
    $\delta=1,\gamma=2$ & lPoisson $\lambda=1$ & 22.33 & 12.85 & 10.00 & 13.00 & 19.00 & 28.00 & 39.00 \\
    $\delta=1,\gamma=2$ & lPoisson $\lambda=3$ & 14.41 & 7.06 &  7.00 &  9.00 & 13.00 & 18.00 & 24.00 \\ \hline
    $\delta=1/2,\gamma=1$ & Fixed $n=1$ & 38.38 & 27.44 & 13.00 & 19.00 & 30.00 & 51.00 & 76.00 \\
    $\delta=1/2,\gamma=1$ & Fixed $n=3$ & 18.73 & 12.98 &  5.00 &  9.00 & 15.00 & 24.00 & 36.00 \\
    $\delta=1/2,\gamma=1$ & lPoisson $\lambda=1$ & 25.92 & 16.44 & 11.00 & 14.00 & 21.00 & 33.00 & 47.00 \\
    $\delta=1/2,\gamma=1$ & lPoisson $\lambda=3$ & 17.78 & 10.74 &  9.00 & 10.00 & 14.00 & 22.00 & 32.00 \\ \hline
    $\delta=1,\gamma=1/2$ & Fixed $n=1$ & 28.15 & 24.23 &  8.00 & 10.00 & 21.00 & 37.00 & 57.00 \\
    $\delta=1,\gamma=1/2$ & Fixed $n=3$ & 16.30 & 12.56 &  4.00 &  7.00 & 13.00 & 22.00 & 33.00 \\
    $\delta=1,\gamma=1/2$ & lPoisson $\lambda=1$ & 19.97 & 16.39 &  5.00 &  8.00 & 15.00 & 25.00 & 42.00 \\
    $\delta=1,\gamma=1/2$ & lPoisson $\lambda=3$ & 14.02 & 10.24 &  4.00 &  7.00 & 11.00 & 19.00 & 27.00 \\ \hline
  \end{tabular}
 \caption{Monte Carlo simulation study results for run length in the out-of-calibration setting. First column shows the data generating setting and alternative specification for the chart and limits. Second column describes the distribution of possible events per time point. lPoisson refers to a Poisson random variable shifted up by 1 to guarantee at least one observation per time point. Remaining columns provide average, standard deviation, and various quantiles of run length.}
  \label{tab:OOC}
\end{table}

\section{Image classification case study}
\label{sec:case_study}

We illustrate the utility of the calibration CUSUM chart using an image classification example. We consider a situation where an image classifier aims to distinguish between vehicles and animals. Perhaps a nature preserve is being photographically monitored and the stewards of the preserve need to recognize the presence of any illicit vehicular activity, regardless of vehicle subtype.

We use the widely available CIFAR-10 data \parencite{Krizhevsky2009cifar}. The CIFAR-10 data includes 60,000 images. The CIFAR-10 dataset contains 10 classes we refer to as subtypes, where vehicle subtypes include airplanes, cars, ships, and trucks, and animal subtypes include birds, cats, deer, dogs, frogs, and horses. Each image is in color and has a 32x32 pixel resolution, and there are 6,000 images of each subtype. 


We engineer two scenarios in the context of this case study (Table \ref{tab:scenario_laoyout}). In Scenario 1, the CNN is fit using the full CIFAR-10 training set containing 50,000 images to produce the $x$ confidence scores. \textcolor{black}{Then, 2,500 of remaining 10,000 $x$ values are randomly selected and used to learn the recalibration mapping (second bubble in Figure \ref{fig:fig2}). Finally, the recalibrated predictions $p$ are obtained for the remaining 7,500 test set predictions which are then subjected to our calibration CUSUM monitoring routine.} To ensure representativeness, the data order was randomized. We hypothesize that the $x$ predictions will be uncalibrated and thus the CUSUM chart will grow rapidly, while the well calibrated $p$ predictions remain calibrated according to the DPCLs. All charts and DPCLs use $\delta_a=1$ and $\gamma_a=1/2$ corresponding the notion that CNN predictions are too spread out as seen in Figure \ref{fig:fig3}.

The second scenario is designed to reflect a loss of calibration during the monitoring phase. \textcolor{black}{We accomplish this by withholding flying subtypes (i.e., birds and planes) from both the CNN training (which produces $x$ predictions) and recalibration exercise (which produces $p$ predictions). We then allow the flying subtypes of objects to appear in the monitoring phase. Since the CNN model is not trained or recalibrated with access to the flying subtypes, we anticipate predictions on these subtypes will be badly calibrated. With respect to the CNN confidence scores $x$, we anticipate the chart will initially grow quickly as we have already shown CNN confidence scores are not well calibrated by default. With respect to the recalibrated $p$ predictions, we hypothesize that the calibration CUSUM chart will respond by maintaining calibration until the flying objects appear in the field, at which point the chart will rapidly grow and signal shortly thereafter.}


We justify excluding fliers from both training and recalibration phases to simulate a scenario where completely new object subtypes enter the field for which the original modeling process is naive of in the first place. If the subtypes were known about during training any sensible analyst would include them in recalibration. If the novel subtypes were discovered during recalibration, the analyst would be aware their trained image classifier would be deficient with respect to these and the model would not be deployed in a monitoring context in the first place. So it is realistic that the new subtypes issue would not manifest in a monitoring problem until the monitoring phase.

\begin{figure}
\centering
\includegraphics[width=1\linewidth]{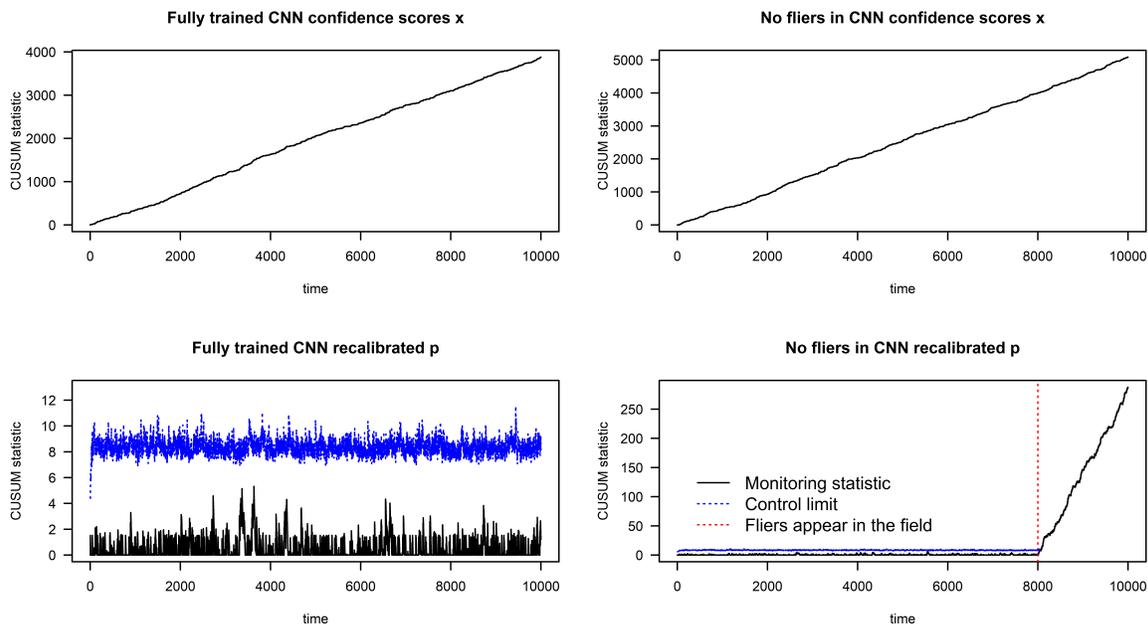}
\caption{Calibration CUSUM charts for the CIFAR-10 vehicle classification monitoring problem using $\delta_a=1$ and $\gamma_a =1/2$. Top left panel shows CNN confidence scores $x$ from a neural net using all training data but with no recalibration. The CUSUM statistic increases rapidly as raw CNN confidence scores are not calibrated. Bottom left shows CNN recalibrated probabilities $p$. \textcolor{black}{Note 2,500 observations were used as a calibration training set so this chart runs from time 1 to 7,500.} The blue line shows the DPCLs generated with $\alpha=10^{-5}$ and $Q=100,000$. Predictions remain calibrated across the time course. Top right shows CNN confidence scores $x$ using CNN that had no fliers (birds or planes) in the training set, and was not recalibrated. Bottom right uses CNN and recalibration that both omit birds and planes. \textcolor{black}{This plot also uses 2,500 points as a calibration set. The first 3,500 monitored time points include only non-flying objects and the chart does not signal. The final 4,000 are randomly ordered with 50\% flying objects (birds and planes) mixed with non-flying objects.} The chart signals at time 3,533.}
\label{fig:fig4}
\end{figure}

\begin{table}
    \begin{tabularx}{\textwidth}{CCCC}
        \hline
        Label & Training set for CNN & Recalibration set includes fliers? & Novel subtypes emerge during monitoring phase? \\
        \hline
        Scenario 1 - Calibration maintained & Full training set, 50,000 images, fliers (i.e., birds, planes) included & Yes & No. All object subtypes accounted for in training and recalibration \\
        \hline
        Scenario 2 - Calibration lost when new subtypes enter monitoring phase & Partial training set, 40,000 images, fliers excluded. & No & Yes. Fliers enter monitoring phase at time $t=3,501$. \\
        \hline
    \end{tabularx}
  \caption{Specific conditions for Scenarios 1 and 2 in the case study.}
  \label{tab:scenario_laoyout}
\end{table}

Figure \ref{fig:fig4} shows four calibration CUSUM charts. The top row includes calibration CUSUM charts for the CNN confidence scores $x$ for scenario 1 (top left) and scenario 2 (top right). In both cases the CUSUM statistic $S_t$ grows without bound because the CNN confidence score is not well calibrated by default. The bottom row of Figure \ref{fig:fig4} shows the $p$ predictions for Scenario 1 (bottom left) and Scenario 2 (bottom right), along with DPCLs in blue. These results indicate that when the CNN was trained and recalibrated including all object subtypes, predictions remained calibrated across the monitoring time course presented here.

The bottom right panel of Figure \ref{fig:fig4} shows the CUSUM chart on the $p$ predictions from Scenario 2. During the first 3,500 time points, non-fliers are sampled and the calibration CUSUM chart does not signal. At time 3,501, the fliers begin appearing in the monitoring phase. The chart signals at time 3,533.

These results demonstrate the ability of the calibration CUSUM chart to avoid signaling when incoming predictions are well calibrated but to signal once calibration is lost for image classification or other binary prediction problems. In this case the emergence of novel subtypes lead to the loss of calibration.

\begin{figure}
\centering
\includegraphics[width=1\linewidth]{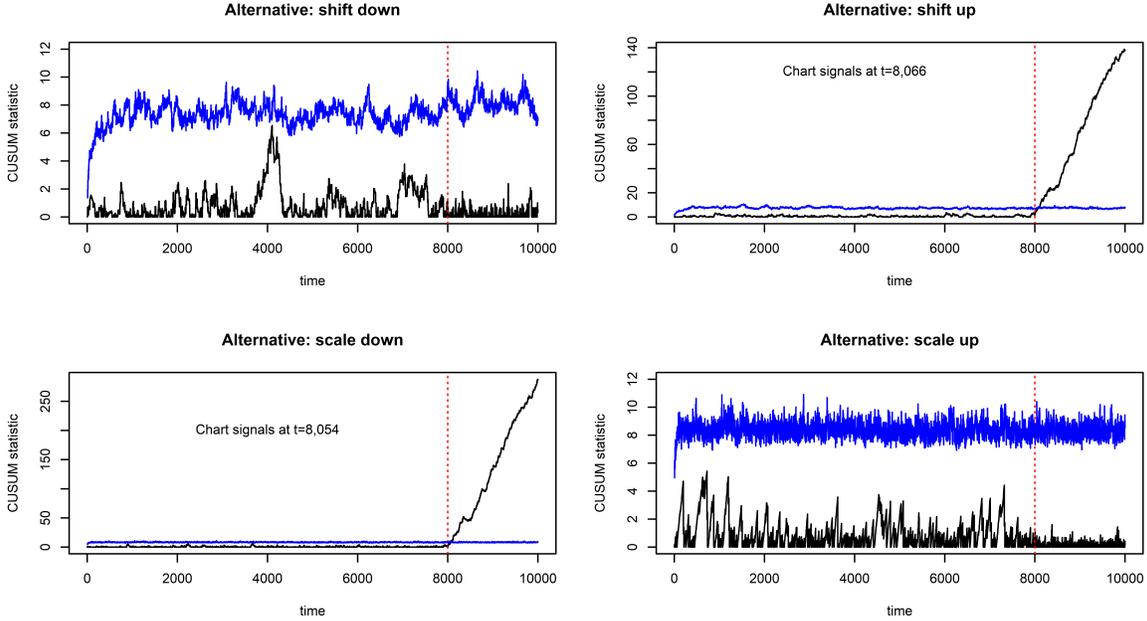}
\caption{\label{fig:fig5}
Calibration CUSUM charts for the CIFAR-10 analysis considering four separate alternatives. The ``shift down'' chart (top left) uses $\delta_a=1/2$, $\gamma_a=1$. The ``shift up'' chart (top right) uses $\delta_a=2$, $ \gamma_a=1$. The ``scale down'' chart (bottom left) uses $\delta_a=1$, $ \gamma_a=1/2$. The ``scale up'' chart (bottom right) uses $\delta_a=1$, $ \gamma_a=2$.}
\end{figure}

Figure \ref{fig:fig5} shows four one-sided charts for the CNN monitoring case study example Scenario 2. These charts show that alternative values corresponding to a shift up or a scale down are able to detect the loss of calibration while the shift down and scale up alternative specifications are not able to detect the loss of calibration. This emphasizes the importance of specifying departures from calibration that correspond to the problem being studied, and/or considering multiple charts simultaneously if the nature of calibration loss is not known in advance.

\section{Discussion}
\label{sec:discussion}

We have reviewed the basic concept of calibration for binary event predictions and developed a calibration CUSUM chart that can detect when the rate of observed events no longer corresponds to available predictions. The techniques have been illustrated using a machine learning approach to image classification. The case study and simulation study demonstrate the effectiveness of the method.

Importantly, the calibration CUSUM chart only requires probability predictions and event data. It does not require the user to have under-the-hood access to the machine learning or statistical models used to make the predictions. Thus, it can be deployed in a wide variety of settings without requiring highly situation-specific technical implementation. The same approach we employ here for image classification could be used much more generally in any classification-based monitoring situation where predictions and observed data are able to be monitored.

A natural question is: what do you do when the calibration CUSUM chart signals in practice? It would be simple enough to recalibrate predictions yet again (i.e., re-deploy the recalibration exercise in the second-to-left box in Figure \ref{fig:fig2}) and continue monitoring on the re-recalibrated predictions. We do not advocate this approach, however, as it is likely important in the context of whatever domain is being monitored to understand why predictions no longer hold with observed event rates. This could be due to new versions of a predictive model being implemented, a model being deployed in a new data-generating situation for which it was not trained, or as we have illustrated here, the emergence of new object subtypes for which the original model was neither trained nor recalibrated to detect. Other domain specific reasons for calibration loss surely exist as well.

The work we present here is not without limitations. Our simulation study did not explore mischaracterization of alternative values - the charts were illustrated using alternative corresponding to the true loss of calibration. In practice, the analyst must specify alternative values on the basis of relevant domain considerations, i.e., consider how much of a shift or scaling on the log odds would be pertinent for detection. Nonetheless, when using our data-agnostic ``shift down'' alternative value of $\gamma_a=1/2$ our chart signaled in Scenario 2 of the case study, which reassures us that reasonable alternative values can be used for the purpose of monitoring using this approach.

Also, in an effort to illustrate clearly how the calibration CUSUM chart behaves under calibration, we used the monitoring phase data to learn the recalibration so as to guarantee that the chart begins in a calibrated state. In practice it seems likely that an additional calibration data set may be needed as the predictions being monitored would not be available to estimate the recalibration parameters $\delta$ and $\gamma$. 

We exclusively used the LLO function for calibration monitoring in this work due to its flexibility, simplicity, and overall effectiveness. This is not the only potential calibration function. For example, \textcite{Prelec1998probability} proposes a two parameter recalibration model that conceptually could be used in place of LLO here. Our experience with the Prelec function is that it is similarly flexible to LLO, as both of these are smooth, two parameter functions that rescale one dimensional probabilities. We anticipate they would perform similarly for the proposed calibration CUSUM approach. See \textcite{Guthrie2024boldness} for a demonstration of Prelec recalibration and comparison with LLO recalibration.

\textcolor{black}{This work has focused on monitoring calibration in the binary classification problem. A natural extension to the multicategory problem might involve using the recently-proposed multicategory linear log odds likelihood function \parencite{Vennos2026multiclass} in the CUSUM chart  to accommodate more than two categories in the outcome being monitored.}

Finally, in Scenario 2, the monitoring phase was structured such that only non-flier observations were presented initially, with flier observations introduced only after all non-fliers had been observed. This clearly demonstrated that the chart does not signal while the novel flier subtype is missing, and does signal when the fliers appear. In practice, it may be more likely that novel subtypes begin to appear in the monitoring data intermittently, so it may take the chart longer to signal depending on the rate at which the novel subtype appears among those subtypes accounted for in the original model training and recalibration exercises.

Despite these limitations, the calibration CUSUM chart is a flexible and easy-to-use approach for the purpose of monitoring prospective predictions and event data. It can enable users to determine if and when model predictions no longer correspond to the rate of observed events, which surely could negatively impact whatever enterprise is of interest.

\section*{Acknowledgments}

This work was supported by the U.S. Department of Defense (DoD) through the Office of the Under Secretary of Defense for Acquisition and Sustainment (OUSD(A\&S)) and the Office of the Under Secretary of Defense for Research and Engineering (OUSD(R\&E)) under Contract HQ003424D0023. The Acquisition Innovation Research Center (AIRC) is a multi-university partnership led and managed by the Stevens Institute of Technology through the Systems Engineering Research Center (SERC) – a federally funded University Affiliated Research Center. Any views, opinions, findings and conclusions or recommendations expressed in this material are those of the authors and do not necessarily reflect the views of the United States Government (including the DoD and any government personnel).

\section*{Data availability statement}

The data used in this manuscript will be made available in the journal's repository upon acceptance and publication.

\section*{AI statement}

We used GPT4 for Python coding assistance to fit the convolutional neural networks used in this work.

\section*{Disclosure of interests}

The authors report no conflicts of interest related to this work.

\section*{\textcolor{black}{Appendix}}

\textcolor{black}{The proposed calibration CUSUM chart generalized the control chart of \textcite{Steiner2000monitoring} in two specific ways. First, the proposed chart accommodates multiple trials per time point while the Steiner approach focuses on a single event per time point. Second, the Steiner chart is a specific case of the proposed CUSUM chart where $\gamma=1$. We demonstrate this result below.}

\textcolor{black}{Following Steiner, we presume one trial per time point. Thus}

$$f_c(y_t)=p_t^{y_t}(1-p_t)^{y_t}$$ 
\textcolor{black}{and}
$$f_u(y_t;\delta,\gamma)= g(p_t;\delta,\gamma)^{y_t}[1-g(p_t;\delta,\gamma)]^{1-y_t}$$

\textcolor{black}{Next fix $\gamma_a=1$. After simplification this implies}

\[
W_t = \text{log}\Big[ \frac{f_u(y_t;\delta_a,\gamma_a=1)}{f_c(y_t)} \Big] = 
\begin{cases}
\text{log}\Big[ \frac{\delta_a}{\delta_a p_t + 1 - p_t}\Big] & \text{ if } y_t =1, \\
\text{log}\Big[ \frac{1}{\delta_a p_t + 1 - p_t}\Big] & \text{ if } y_t =0.
\end{cases}
\quad
\]

\textcolor{black}{This matches \textcite{Steiner2000monitoring} Equation (2.3) when the null odds ratio $R_0=1$ and the alternative odds ratio $R_A=\delta_a$.}


\printbibliography


\end{document}